\documentclass{article}

\usepackage{microtype}
\usepackage{graphicx}
\usepackage{subfigure}
\usepackage{booktabs} 
\usepackage[table]{xcolor}
\usepackage{multirow}

\usepackage{hyperref}

\usepackage[preprint]{icml2026}

\usepackage{amsmath}
\usepackage{amssymb}
\usepackage{mathtools}
\usepackage{amsthm}

\usepackage{amsmath,amsfonts,bm}

\def\eqref#1{equation~\ref{#1}}

\def\1{\bm{1}}

\DeclareMathAlphabet{\mathsfit}{\encodingdefault}{\sfdefault}{m}{sl}
\SetMathAlphabet{\mathsfit}{bold}{\encodingdefault}{\sfdefault}{bx}{n}

\usepackage{url}
\usepackage{algorithm}
\usepackage{algorithmic}

\definecolor{Gray}{gray}{0.9}

\usepackage[capitalize,noabbrev]{cleveref}

\theoremstyle{plain}

\theoremstyle{definition}

\theoremstyle{remark}

\icmltitlerunning{LVRPO: Language-Visual Alignment with GRPO for Multimodal Understanding and Generation}

\begin{document}

\twocolumn[
\icmltitle{LVRPO: Language-Visual Alignment with GRPO for Multimodal \\ Understanding and Generation}




\begin{icmlauthorlist}
\icmlauthor{Shentong Mo}{yyy,comp}
\icmlauthor{Sukmin Yun}{sch}
\end{icmlauthorlist}

\icmlaffiliation{yyy}{Department of Machine Learning, CMU, USA}
\icmlaffiliation{comp}{Department of Machine Learning, MBZUAI, UAE}
\icmlaffiliation{sch}{Department of Artificial Intelligence, Hanyang University ERICA, South Korea}

\icmlcorrespondingauthor{Sukmin Yun}{sukminyun@hanyang.ac.kr}

\icmlkeywords{Machine Learning, ICML}

\vskip 0.3in
]



\printAffiliationsAndNotice{}  

\begin{abstract}

Unified multimodal pretraining has emerged as a promising paradigm for jointly modeling language and vision within a single foundation model. 
However, existing approaches largely rely on implicit or indirect alignment signals and remain suboptimal for simultaneously supporting multimodal understanding and generation, particularly in settings that require fine-grained language–visual reasoning and controllable generation.
In this work, we propose LVRPO, a language-visual reinforcement-based preference optimization framework that explicitly aligns language and visual representations using Group Relative Policy Optimization (GRPO). 
Instead of introducing additional alignment losses at the representation level, LVRPO directly optimizes multimodal model behaviors through preference-driven reinforcement signals, encouraging consistent and semantically grounded interactions between language and vision across both understanding and generation tasks. This formulation enables effective alignment without requiring auxiliary encoders or handcrafted cross-modal objectives, and naturally extends to diverse multimodal capabilities.
Empirically, LVRPO consistently outperforms strong unified-pretraining baselines on a broad suite of benchmarks spanning multimodal understanding, generation, and reasoning. 

\end{abstract}

\section{Introduction}

A unified foundation model capable of seamlessly bridging multimodal understanding and generation has led to the emergence of unified pretraining as a dominant paradigm. Unlike earlier disjointed pipelines~\cite{mo2024dmtjepa,mo2026gmail}, these models aim to model vision within a single architecture, often utilizing large-scale data to foster emergent uni-modal capabilities. 
Recent breakthroughs, such as BAGEL~\cite{deng2025bagel}, have demonstrated that scale alone can induce impressive multimodal reasoning and generation without requiring explicit alignment objectives. Simultaneously, methods like REPA~\cite{yu2025repa} have shown that aligning generative features with non-generative teachers (\textit{e.g.}, DINOv2~\cite{oquab2024dinov2}) can significantly accelerate convergence for vision-centric tasks.

Despite these advances, a fundamental gap remains: existing approaches largely rely on implicit or indirect alignment signals. While BAGEL relies on the sheer density of interleaved tokens to learn associations, it often struggles with fine-grained semantic grounding, leading to hallucinations or failures in complex instruction following. On the other hand, representation-level alignment methods like REPA, which typically rely on a fixed visual encoder to provide a "semantic anchor," are often suboptimal for tasks requiring bidirectional synergy between language and vision. These methods typically prioritize vision-centric denoising over the nuanced, instruction-driven reasoning required for controllable generation and world-knowledge-integrated understanding.

The core challenge in unified multimodal modeling lies in the discrepancy between internal representation alignment and end-to-end task performance. Existing methodologies generally face three critical hurdles:
(i) Representation-based alignment, such as REPA~\cite{yu2025repa}’s use of DINOv2~\cite{oquab2024dinov2} features, focuses on minimizing the distance between the model’s latent states and a frozen visual teacher. However, visual encoders optimized for discriminative tasks often capture different features than those required for high-fidelity generation or complex reasoning. Aligning to these features provides a "hint" but does not necessarily translate to compositional understanding or the ability to follow intricate textual constraints in generation.
(ii) In unified models, the objective functions for understanding (often cross-entropy) and generation (often diffusion-based or autoregressive) can conflict. Indirect signals struggle to balance these: a model might excel at captioning an image but fail to edit that same image based on a nuanced instruction, as the underlying "world knowledge" is not consistently activated across both heads.
(iii) Current paradigms lack a mechanism to penalize "semantically incorrect but statistically plausible" outputs. For instance, if a model generates a "blue car" when asked for a "red car," traditional reconstruction losses provide only a pixel-level or token-level penalty. There is no explicit signal that reinforces the logical preference for the correct attribute binding, leading to suboptimal performance on benchmarks like GenEval~\cite{ghosh2023geneval} and WISE~\cite{niu2025wise} where precision is paramount.

In this work, we introduce LVRPO, a framework for Language-Visual Reinforcement-based Preference Optimization that explicitly aligns language and visual representations through the lens of behavior-driven reinforcement. Unlike traditional methods that introduce auxiliary alignment losses at the feature level, LVRPO leverages Group Relative Policy Optimization (GRPO)~\cite{shao2024deepseekmath} to directly optimize the model's responses. 
By treating multimodal consistency as a preference-driven signal, LVRPO encourages the model to generate outputs, whether text or images, that are semantically grounded and instruction-faithful.
While REPA aligns internal hidden states to a static vision teacher, LVRPO optimizes the end-to-end behavior. We utilize the SigLIP 2~\cite{tschannen2025siglip} encoder's dense semantic features not as a rigid target for distillation, but as a robust foundation for evaluating and reinforcing multimodal consistency during preference optimization.
Rather than relying on emergent alignment from scale in BAGEL, LVRPO introduces an explicit preference stage that "closes the loop" between understanding and generation, leading to superior grounding in complex scenarios.

Our empirical evaluation demonstrates the efficacy of LVRPO across a comprehensive suite of benchmarks. On multimodal understanding tasks, LVRPO consistently outperforms strong unified baselines like BAGEL. In the realm of generation, it achieves state-of-the-art results on GenEval, particularly in compositional accuracy. Notably, LVRPO shows a significant leap in world-knowledge-based reasoning on WISE, suggesting that preference optimization effectively bridges the gap between linguistic knowledge and visual realization. Finally, on image editing benchmarks such as GEdit-Bench and the reasoning-heavy IntelligentBench, LVRPO exhibits superior instruction-following, enabling precise and semantically aware visual manipulation that surpasses existing open-source and many proprietary competitors.

Our contributions can be summarized as follows:
\begin{itemize}
    \item We propose a novel language-visual reinforcement-based preference optimization framework that explicitly aligns understanding and generation through behavioral signals rather than static representation mimicry.
    \item We extend Group Relative Policy Optimization to the multimodal domain, enabling efficient, critic-free alignment of a unified MoT backbone. 
    \item We provide theoretical justifications for the synergy between reasoning and generation in unified models, proving that LVRPO maximizes a lower bound on cross-modal mutual information while maintaining training stability through gradient decoupling.
    \item We demonstrate the empirical superiority of LVRPO across a comprehensive suite of benchmarks.
\end{itemize}

\section{Related Work}

\noindent\textbf{Unified Multimodal Foundation Models.}
Recent research has shifted from modular pipelines toward unified architectures that model language and vision within a single transformer backbone. Early attempts like Chameleon~\cite{chameleon} and Show-o~\cite{show-o} demonstrated the feasibility of interleaved tokenization for discrete multimodal modeling. More recently, BAGEL~\cite{deng2025bagel} and Emu3~\cite{emu3} have advanced this paradigm by utilizing large-scale pretraining to induce emergent reasoning and high-fidelity generation. However, these models primarily rely on implicit alignment through next-token prediction or flow matching. LVRPO builds upon this unified foundation by introducing an explicit behavioral alignment stage, ensuring that emergent capabilities are grounded in semantically consistent interactions.

\noindent\textbf{Multimodal Representation Alignment.}
A common strategy to bridge the gap between modalities is to align the model’s internal representations with a pretrained teacher. REPA~\cite{yu2025repa} utilizes denoising feature alignment to bridge generative models with discriminative teachers like DINOv2~\cite{oquab2024dinov2}. Similarly, methods like VILA-U~\cite{vila-u} and Janus~\cite{janus2024} focus on cross-modal bottleneck layers or shared latent spaces. While effective for feature transfer, these methods often suffer from \textit{feature rigidness}, forcing the model to inherit the inductive biases of the teacher. LVRPO departs from this by shifting from representation alignment (mimicking hidden states) to behavioral alignment (optimizing outputs), utilizing SigLIP 2~\cite{tschannen2025siglip} as a semantic referee rather than a distillation target.

\noindent\textbf{Preference Optimization in Multimodal Learning.}
Reinforcement Learning from Human Feedback (RLHF) and Direct Preference Optimization (DPO) have revolutionized the alignment of Large Language Models (LLMs) with human intent. In the multimodal domain, recent works have explored using DPO for text-to-image models to improve aesthetic quality or instruction following~\cite{wallace2023diffusion}. However, applying these to unified models is challenging due to the high computational cost of the Critic network in standard PPO. We leverage Group Relative Policy Optimization (GRPO)~\cite{shao2024deepseekmath}, which eliminates the need for a separate value function by using group-relative rewards. By extending GRPO to unified multimodal tasks, LVRPO enables efficient alignment of both understanding and generation pathways simultaneously, addressing the ``seesaw effect'' often observed in multi-task learning.

\section{Method}

In this section, we present LVRPO, a framework designed to explicitly align language and visual modalities via reinforcement-based preference optimization. Unlike previous methods that rely on static representation distillation or implicit alignment through large-scale pretraining, LVRPO treats the unified multimodal model as a policy that is optimized based on the semantic consistency and instruction-following quality of its outputs.

\begin{figure*}[t]
\centering
\includegraphics[width=0.98\linewidth]{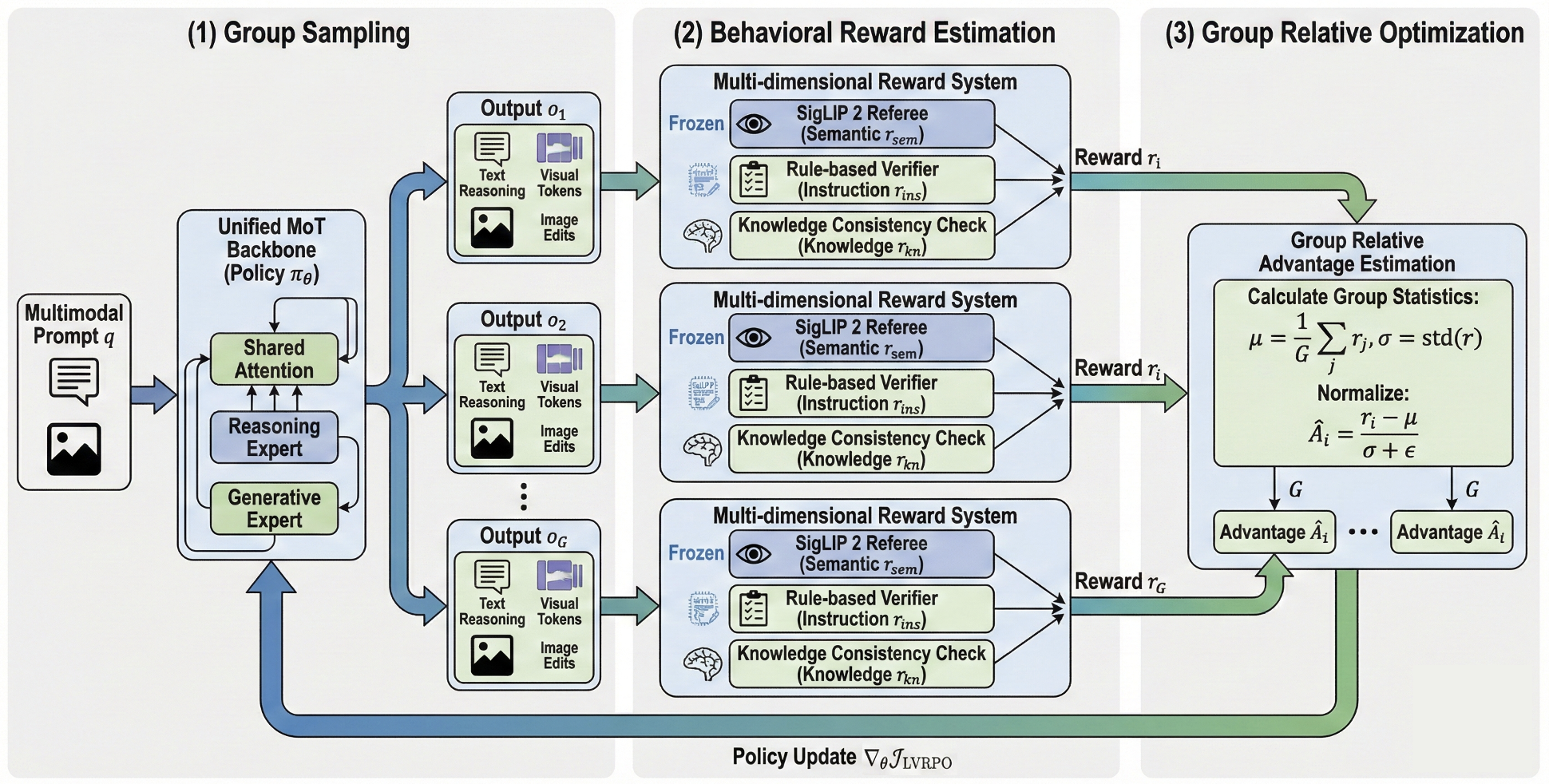}
\vspace{-0.5em}
\caption{Illustration of the proposed LVRPO framework for unified multimodal understanding and generation. 
The framework proceeds in three stages: 
(1) Group Sampling: For a given multimodal prompt $q$, the unified \textit{MoT backbone} (initialized from BAGEL) samples a group of $G$ independent outputs $\{o_1, \dots, o_G\}$, which can include text reasoning, visual tokens, or image edits. 
(2) Behavioral Reward Estimation: Each output is evaluated by a multi-dimensional reward system. We utilize a frozen \textit{SigLIP 2} referee to provide dense semantic grounding signals ($r_{sem}$), alongside rule-based verifiers for instruction following ($r_{ins}$) and knowledge consistency ($r_{kn}$). 
(3) Group Relative Optimization: Instead of using a separate critic network, LVRPO employs \textit{Group Relative Policy Optimization} to compute the advantage $\hat{A}_i$ for each sample by normalizing its reward against the mean and standard deviation of the group. This behavioral feedback $\nabla_\theta \mathcal{J}_{\mathrm{LVRPO}}$ is backpropagated through the shared attention layers to jointly optimize the reasoning and generative experts, enforcing cross-modal consistency without the need for auxiliary representation-level alignment losses.
}
\label{fig:main_image}
\vspace{-1.0em}
\end{figure*}

\subsection{Preliminaries}

\textbf{Unified Multimodal Modeling.} We consider a unified transformer-based architecture $\mathcal{M}$ parameterized by $\theta$. The model processes interleaved sequences of text tokens $\mathbf{T} = \{t_1, \dots, t_n\}$ and visual patches $\mathbf{V} = \{v_1, \dots, v_m\}$. The model is trained to minimize a joint objective, typically combining autoregressive language modeling and latent visual reconstruction:
\begin{equation}
    \mathcal{L}_{unified} = \mathcal{L}_{text}(\mathbf{T} | \mathbf{V}) + \lambda \mathcal{L}_{vis}(\mathbf{V} | \mathbf{T}).
\end{equation}

\textbf{GRPO.} To avoid the computational burden of a separate Critic network in traditional RLHF, LVRPO adopts the GRPO objective. For a given prompt $q$, we sample a group of $G$ outputs $\{o_1, o_2, \dots, o_G\}$ from the current policy $\pi_{\theta}$. The objective is defined as:
\begin{equation}
\begin{aligned}
\mathcal{J}_{\mathrm{GRPO}}(\theta)
&= \mathbb{E}_{q \sim P(q),\, \{o_i\} \sim \pi_{\theta}}
\Bigg[
\frac{1}{G} \sum_{i=1}^{G}
\mathcal{L}_{\mathrm{CLIP}}(o_i, q)\, \hat{A}_i \\
&\qquad
- \beta \, \mathbb{D}_{\mathrm{KL}}\!\left(
\pi_\theta \,\|\, \pi_{\mathrm{ref}}
\right)
\Bigg].
\end{aligned}
\end{equation}
where $\mathcal{L}_{CLIP}$ denotes the surrogate loss used in PPO, and $\hat{A}_i$ is the relative advantage of output $o_i$ within its group, computed as:
\begin{equation}
    \hat{A}_i = \frac{r_i - \text{mean}(\{r_1, \dots, r_G\})}{\text{std}(\{r_1, \dots, r_G\})}.
\end{equation}
Here, $r_i$ represents the multi-dimensional reward assigned to the $i$-th sample.

\subsection{LVRPO: Behavioral Multimodal Alignment}

Current state-of-the-art methods like REPA~\cite{yu2025repa} utilize a representation-level alignment objective, typically formulated as a mean-squared error (MSE) between model features $h_\theta$ and a teacher's features $\Phi_{tea}$:
\begin{equation}
    \mathcal{L}_{align} = \mathbb{E}_{v \sim \mathcal{D}} [ \| f(h_\theta(v)) - \Phi_{tea}(v) \|_2^2 ].
\end{equation}
While effective for vision-centric tasks, we argue this is suboptimal for \textit{unified} multimodal models for two reasons: (1) Feature Rigidness: It forces the model to mimic the teacher's inductive biases (\textit{e.g.}, DINOv2's patch-level focus), which may conflict with the generative requirements of the language-vision head. (2) Objective Mismatch: Minimizing latent distance does not guarantee that the model's \textit{behavior} (\textit{e.g.}, generating a specific object) follows the instruction. LVRPO motivates alignment through the lens of expected utility maximization, where the model is rewarded for the semantic correctness of its output rather than the configuration of its hidden states.

\subsubsection{Reward Formulation and Advantage Estimation}
In LVRPO, we define a multimodal reward function $\mathcal{R}(o, q)$ that evaluates a generated output $o$ given a prompt $q$. Unlike traditional PPO, we utilize Group Relative Policy Optimization (GRPO)~\cite{shao2024deepseekmath} to derive the advantage $\hat{A}_i$ without a critic network.

For a group of $G$ outputs $\{o_i\}_{i=1}^G$, the reward $r_{sem}$ for each sample is a composition of semantic and structural signals:
\begin{equation}
    r_i = \underbrace{\lambda_1 \cdot \text{sim}(\Phi_{sig}(V_i), \Psi_{sig}(T_q))}_{\text{Semantic Grounding}} + \underbrace{\lambda_2 \cdot \mathbb{I}(\text{rules}(o_i, q))}_{\text{Logical Constraints}}
\end{equation}
where $\mathbb{I}(\cdot)$ is an indicator function for verifiable constraints (\textit{e.g.}, image resolution, object counts, or text formatting). The advantage is then computed via group-relative normalization:
\begin{equation}
    \hat{A}_i = \frac{r_i - \frac{1}{G}\sum_{j=1}^G r_j}{\sigma(\{r_j\}_{j=1}^G) + \epsilon}
\end{equation}

\noindent\textbf{Instruction Following ($r_{ins}$).} We define $r_{ins}$ as a binary satisfaction score over a set of verifiable rules $\mathcal{C}_q$ derived from the prompt (e.g., aspect ratio, object count, text presence):
\begin{equation}
    r_{ins}(o, q) = \frac{1}{|\mathcal{C}_q|} \sum_{c \in \mathcal{C}_q} \mathbb{I}(o \text{ satisfies } c).
\end{equation}

\noindent\textbf{Knowledge Consistency ($r_{kn}$).} To ensure factual grounding, we employ a Visual Question Answering (VQA) proxy (\textit{e.g.}, PaLI-3) to verify specific knowledge claims. For a prompt requiring factual attributes (\textit{e.g.}, "A nebulous star forming region"), the reward is:
\begin{equation}
\begin{aligned}
    r_{kn}(o, q) & = \text{Confidence}( \text{VQA}(o, \text{"Is this consistent with } \\
    & q_{fact} \text{?"}) == \text{"Yes"} ).
\end{aligned}
\end{equation}

\subsubsection{KL-Regularized Policy Improvement}
We provide a derivation showing that the LVRPO objective effectively moves the unified model toward an optimal multimodal distribution. 

\textbf{Proposition 1.} \textit{The GRPO update rule maximizes a lower bound on the expected reward while maintaining a trust region around the reference multimodal policy $\pi_{ref}$.}

\textbf{Proof Sketch.}
Consider the generalized RL objective with a KL-divergence constraint:
\begin{equation}
    \max_{\pi_\theta} \mathbb{E}_{o \sim \pi_\theta} [\mathcal{R}(o, q)] - \beta \mathbb{D}_{KL}(\pi_\theta \| \pi_{ref})
\end{equation}
Taking the derivative with respect to $\theta$ and applying the log-derivative trick, we obtain:
\begin{equation}
    \nabla_\theta \mathcal{J} = \mathbb{E}_{o \sim \pi_\theta} [(\mathcal{R}(o, q) - \beta \log \frac{\pi_\theta(o|q)}{\pi_{ref}(o|q)}) \nabla_\theta \log \pi_\theta(o|q)]
\end{equation}
In the GRPO framework, the baseline is implicitly defined by the group mean $\bar{r} = \frac{1}{G}\sum r_j$. By substituting $\hat{A}_i \approx \mathcal{R}(o, q) - \bar{r}$, we arrive at the LVRPO update rule:
\begin{equation}
\begin{aligned}
\nabla_\theta \mathcal{J}_{\mathrm{LVRPO}}
&\approx \frac{1}{G} \sum_{i=1}^{G}
\Big(
\hat{A}_i \nabla_\theta \log \pi_\theta(o_i \mid q) \\
&\qquad
- \beta \, \nabla_\theta
\mathbb{D}_{\mathrm{KL}}\!\left(
\pi_\theta \,\|\, \pi_{\mathrm{ref}}
\right)
\Big)
\end{aligned}
\end{equation}
This demonstrates that LVRPO optimizes for the relative preference of multimodal behaviors. By using the group mean as a baseline, LVRPO filters out common multimodal noise (\textit{e.g.}, generic image features) and focuses the gradient signal on the specific components of the output that satisfy the language-visual grounding constraints.

\subsubsection{Cross-Modal Grounding with SigLIP 2}
To ensure the reward $r_i$ is fine-grained, we do not simply take the global cosine similarity. We define a \textit{Dense Grounding Reward}:
\begin{equation}
    r_{dense} = \frac{1}{|K|} \sum_{k \in K} \max_{p \in \text{patches}} (\phi_{sig}(v_p) \cdot \psi_{sig}(t_k))
\end{equation}
where $K$ is the set of key semantic tokens in the prompt. This forces the model to achieve \textit{spatial-semantic alignment}, ensuring that specific words in the instruction are grounded in corresponding regions of the generated visual manifold.

\begin{table*}[!htb]
    \centering
    \setlength{\tabcolsep}{4pt}
    \renewcommand{\arraystretch}{1.2}
    \caption{Comparison of text-to-image generation on GenEval benchmark.
    }
    \label{tab:geneval}
    \vspace{-0.5em}
    \scalebox{0.75}{
    \begin{tabular}{clccccccc}
        \toprule
        \textbf{Type} & \textbf{Model}  & \textbf{Single Obj.} & \textbf{Two Obj.} & \textbf{Counting} & \textbf{Colors} & \textbf{Position} & \textbf{Color Attri.} & \textbf{Overall$\uparrow$} \\
        \midrule
        \multirow{8}{*}{\rotatebox{90}{\textit{Gen. Only}}}
        & PixArt-$\alpha$~\cite{chen2024pixart} &  0.98 & 0.50 & 0.44 & 0.80 & 0.08 & 0.07 & 0.48 \\
        & SDv$2.1$~\cite{rombach2022high} & 0.98 & 0.51 & 0.44 & 0.85 & 0.07 & 0.17 & 0.50 \\
        & DALL-E $2$~\cite{dalle2}  & 0.94 & 0.66 & 0.49 & 0.77 & 0.10 & 0.19 & 0.52 \\
        & Emu$3$-Gen ~\cite{emu3}  & 0.98 & 0.71 & 0.34 & 0.81 & 0.17 & 0.21 & 0.54 \\
        & SDXL~\cite{sdxl} &  0.98 & 0.74 & 0.39 & 0.85 & 0.15 & 0.23 & 0.55 \\
        & DALL-E $3$~\cite{dalle3} & 0.96 & 0.87 & 0.47 & 0.83 & 0.43 & 0.45 & 0.67 \\
        & SD3-Medium~\cite{SD3} & 0.99 & 0.94 & 0.72 & 0.89 & 0.33 & 0.60 & 0.74 \\
        & FLUX.1-dev$^{\dagger}$~\cite{flux} & 0.98 & 0.93 & 0.75 & 0.93 & 0.68 & 0.65 & \emph{0.82} \\
        \midrule
        \multirow{13}{*}{\rotatebox{90}{\textit{Unified}}}
        & Chameleon~\cite{chameleon} &  - & - & - & - & - & - & 0.39 \\
        & LWM~\cite{lwm} &  0.93 & 0.41 & 0.46 & 0.79 & 0.09 & 0.15 & 0.47 \\
        & SEED-X~\cite{seed-x}  & 0.97 & 0.58 & 0.26 & 0.80 & 0.19 & 0.14 & 0.49 \\
        & TokenFlow-XL~\cite{qu2024tokenflow} &  0.95 & 0.60 & 0.41 & 0.81 & 0.16 & 0.24 & 0.55 \\
        & ILLUME~\cite{wang2024illume} &  0.99 & 0.86 & 0.45 & 0.71 & 0.39 & 0.28 & 0.61 \\
        & Janus~\cite{janus2024} & 0.97 & 0.68 & 0.30 & 0.84 & 0.46 & 0.42 & 0.61 \\
        & Transfusion~\cite{transfusion} & - & - & - & - & - & - & 0.63 \\
        & Emu$3$-Gen\cite{emu3} & 0.99 & 0.81 & 0.42 & 0.80 & 0.49 & 0.45 & 0.66 \\
        & Show-o~\cite{show-o} &  0.98 & 0.80 & 0.66 & 0.84 & 0.31 & 0.50 & 0.68 \\
        & Janus-Pro-7B~\cite{januspro2025} &  0.99 & 0.89 & 0.59 & 0.90 & 0.79 & 0.66 & 0.80 \\
        & MetaQuery-XL~\cite{pan2025transfer} &  -& - & - & -& -& -& 0.80 \\
        & BAGEL~\cite{deng2025bagel} & 0.99 & 0.94  & 0.81 & 0.88 & 0.64 & 0.63 & 0.82\\
        & LVRPO (ours) & \bf 0.99 & \bf 0.96 & \bf 0.87 & \bf 0.96 & \bf 0.83 & \bf 0.81 & \bf 0.91 \\
    \bottomrule
    \end{tabular}}
    \vspace{-1.0em}
\end{table*}

\subsection{Unified Generation and Understanding}

A fundamental limitation of dual-stream models is the "semantic disconnect" where the reasoning engine (LLM) and the generative engine (Diffusion/Flow) operate in separate latent manifolds. LVRPO leverages BAGEL~\cite{deng2025bagel} as the foundation model, which posits that visual generation should be a downstream consequence of linguistic "thinking." By utilizing a shared attention transformer, we ensure that the generative trajectory is conditioned on the full hidden state of the reasoning process, effectively enabling \textit{Multimodal Chain-of-Thought (M-CoT)}.

We define the unified model as a mapping $\mathcal{F}_\theta$ that operates on a sequence of mixed tokens $\mathbf{S}$. For a visual generation task, we employ Rectified Flow Matching to map a simple noise distribution $p_0$ to the image latent distribution $p_1$. The model learns a velocity field $v_\theta(\mathbf{x}_t, t, \mathbf{c})$ where $\mathbf{c}$ is the multimodal context (text and semantic vision tokens). 

The training objective for the generative pathway is:
\begin{equation}
    \mathcal{L}_{gen}(\theta) = \mathbb{E}_{t \sim [0,1], \mathbf{x}_0, \mathbf{x}_1} \left[ \| v_\theta(\mathbf{x}_t, t, \mathbf{c}) - (\mathbf{x}_1 - \mathbf{x}_0) \|^2 \right]
\end{equation}
where $\mathbf{x}_t = t\mathbf{x}_1 + (1-t)\mathbf{x}_0$ is the linear interpolation between noise $\mathbf{x}_0$ and latent $\mathbf{x}_1$. Simultaneously, the understanding pathway optimizes the standard autoregressive loss:
\begin{equation}
    \mathcal{L}_{und}(\theta) = - \sum_{j} \log P(t_j | \mathbf{t}_{<j}, \mathbf{V}_{sem}).
\end{equation}

To alleviate modality interference, where the high-variance gradients of pixel reconstruction might degrade the discrete reasoning of language, we utilize the MoT architecture. In each layer, a router $\mathcal{G}$ assigns tokens to either a \textit{Reasoning Expert} $\mathcal{E}_{und}$ or a \textit{Generative Expert} $\mathcal{E}_{gen}$:
\begin{equation}
    \text{Output}(\mathbf{s}) = \text{MHA}(\mathbf{s}) + \sum_{i \in \{und, gen\}} \mathcal{G}(\mathbf{s})_i \cdot \mathcal{E}_i(\mathbf{s}).
\end{equation}
This decoupling allows $\mathcal{E}_{gen}$ to specialize in the numerical precision required for flow matching while $\mathcal{E}_{und}$ preserves the world knowledge required for benchmarks like \textit{WISE}.

\textbf{Proposition 2.} \textit{Under the LVRPO framework, the joint optimization of $\mathcal{L}_{gen}$ and $\mathcal{L}_{und}$ converges to a state where the generative velocity field is semantically consistent with the language posterior.}

\textbf{Proof Sketch.} Please See Appendix.

\begin{table*}[!htb]
    \centering
    \caption{Comparison of world knowledge reasoning on WISE.
    }
\label{tab:wisescore}
    \vspace{-0.5em}
    \setlength{\tabcolsep}{8pt}
    \scalebox{0.75}{
    \begin{tabular}{clccccccc}
    \toprule
    \textbf{Type} & \textbf{Model} & \textbf{Cultural}  & \textbf{Time}     & \textbf{Space}    & \textbf{Biology}    & \textbf{Physics} & \textbf{Chemistry} & \textbf{Overall$\uparrow$} \\
    \midrule
            \multirow{6}{*}{\rotatebox{90}{\textit{Gen. Only}}} &
 SDv1.5~\cite{rombach2022high} & 0.34 & 0.35& 0.32&0.28 &0.29 &0.21 & 0.32\\
& SDXL~\cite{sdxl} &0.43  & 0.48 &0.47  &0.44  &0.45 &0.27 & 0.43 \\
& SD3.5-large~\cite{SD3} & 0.44 &0.50 &0.58  & 0.44&0.52 &0.31 & 0.46 \\
& PixArt-Alpha~\cite{chen2024pixart} & 0.45  & 0.50& 0.48 & 0.49& 0.56 &0.34 & 0.47\\
& playground-v2.5~\cite{li2024playground} & 0.49  &0.58  & 0.55&0.43  & 0.48&0.33 & 0.49 \\
& FLUX.1-dev~\cite{flux} & 0.48  &0.58 &0.62  &0.42  &0.51 & 0.35 & 0.50 \\
    \midrule
    \multirow{9}{*}{\rotatebox{90}{\textit{Unified}}} 
& Janus~\cite{janus2024} &0.16 &0.26 &0.35 & 0.28 &0.30 & 0.14& 0.23\\
 &VILA-U~\cite{vila-u} & 0.26 &0.33  & 0.37 &0.35  &0.39 &0.23 & 0.31\\
& Show-o-512~\cite{show-o} & 0.28 &0.40  &0.48 & 0.30& 0.46 & 0.30 & 0.35\\
& Janus-Pro-7B~\cite{januspro2025} & 0.30& 0.37& 0.49 & 0.36&0.42 &0.26 & 0.35 \\
& Emu3~\cite{emu3} & 0.34 & 0.45 & 0.48 & 0.41  & 0.45 & 0.27 & 0.39 \\
& MetaQuery-XL~\cite{pan2025transfer} & 0.56& 0.55 &0.62 &  0.49 &  0.63 & 0.41 & 0.55 \\
& GPT-4o~\cite{openai2025chatgpt4o} &0.81 & 0.71 & 0.89 & 0.83 &0.79 &0.74 & 0.80 \\
& BAGEL~\cite{deng2025bagel} & 0.76 & 0.69 & 0.75 & 0.65 & 0.75 & 0.58 & 0.70 \\
& LVRPO (ours) & \bf 0.86 & \bf 0.78 & \bf 0.93 & \bf 0.89 & \bf 0.87 & \bf 0.79 & \bf 0.85 \\
        \bottomrule
    \end{tabular}}
    \vspace{-1.0em}
\end{table*}

\begin{table*}[!htb]
    \centering
    \caption{Comparison on GEdit-Bench.}
\label{tab:gedit}
\vspace{-0.5em}
\scalebox{0.75}{
\begin{tabular}{cl|ccc|ccc}
\toprule
\multirow{2}{*}{\textbf{Type}} & \multirow{2}{*}{\textbf{Model}} &
\multicolumn{3}{c|}{\textbf{GEdit-Bench-EN (Full set)$\uparrow$}} 
& \multicolumn{3}{c}{\textbf{GEdit-Bench-CN (Full set)$\uparrow$}} \\
\cmidrule(lr){3-8}
& & \textbf{G\_SC} & \textbf{G\_PQ} & \textbf{G\_O} 
& \textbf{G\_SC} & \textbf{G\_PQ} & \textbf{G\_O} \\
\midrule
\multirow{2}{*}{\textit{Private}}&Gemini 2.0~\cite{gemini220250312}                       & 6.73 & 6.61 & 6.32 & 5.43 & 6.78 & 5.36  \\
&GPT-4o~\cite{openai2025chatgpt4o}                         & 7.85 & 7.62 & 7.53 & 7.67 & 7.56 & 7.30  \\
\midrule
\multirow{7}{*}{\textit{Open-source}}
&Instruct-Pix2Pix~\cite{Brooks2022InstructPix2PixLT} & 3.58 & 5.49 & 3.68 & - & - & - \\
&MagicBrush~\cite{zhang2023magicbrush}               & 4.68 & 5.66 & 4.52 &  - & - & - \\
&AnyEdit~\cite{yu2024anyedit}                        & 3.18 & 5.82 & 3.21 &  - & - & - \\
&OmniGen~\cite{xiao2024omnigen}                      & 5.96 & 5.89 & 5.06 & - & - & - \\
&Step1X-Edit~\cite{liu2025step1xeditpracticalframeworkgeneral}& 7.09 & 6.76 & 6.70 & 7.20 & 6.87 & 6.86   \\
& BAGEL~\cite{deng2025bagel}  & 7.36 & 6.83 & 6.52 & 7.34 & 6.85 & 6.50 \\
& LVRPO (ours) & \bf 7.92 & \bf 7.75 & \bf 7.67 & \bf 7.78 & \bf 7.68 & \bf 7.56  \\
\bottomrule
\end{tabular}}
\vspace{-1.0em}
\end{table*}

\begin{table}[!htb]
    \centering
    \caption{Comparison on IntelligentBench.}
\label{tab:IntelligentBench}
\vspace{-0.5em}
    \setlength{\tabcolsep}{8pt}
    \scalebox{0.75}{
    \begin{tabular}{clc}
    \toprule
    \textbf{Type} & \textbf{Model}  & \textbf{Score$\uparrow$} \\
    \midrule
    \multirow{2}{*}{\textit{Private}}
       & GPT-4o~\cite{openai2025chatgpt4o} & 78.9\\
       & Gemini 2.0~\cite{gemini220250312} & 57.6 \\
    \midrule
        \multirow{3}{*}{\textit{Open-source}}& Step1X-Edit~\cite{liu2025step1xeditpracticalframeworkgeneral} & 14.9\\ 
& BAGEL~\cite{deng2025bagel} & 55.3 \\
& LVRPO (ours) & \bf 58.7 \\
        \bottomrule
    \end{tabular}}
    \vspace{-1.0em}
\end{table}

\begin{table*}[!htb]
    \centering
    \setlength{\tabcolsep}{2pt}
    \renewcommand{\arraystretch}{1.2}
    \caption{Comparison with state-of-the-arts on viusal understanding benchmarks. 
    }
    \label{tab: image_understanding}
    \vspace{-0.5em}
    \scalebox{0.78}{
    \begin{tabular}{clccccccccc}
        \toprule
        \textbf{Type} & \textbf{Model} & \textbf{\# LLM Params} & \textbf{MME-P$\uparrow$} & \textbf{MME-S$\uparrow$} &\textbf{MMBench$\uparrow$} & \textbf{MMMU$\uparrow$} & \textbf{MM-Vet$\uparrow$} & \textbf{MathVista$\uparrow$} & \textbf{MMVP$\uparrow$} \\
        \midrule
        \multirow{13}{*}{\rotatebox{90}{\textit{Und. Only}}}
        & InternVL2~\cite{internvl2}& 1.8B & 1440 & 1877 &  73.2  & 34.3 & 44.6 & 46.4 & 35.3 \\
        & InternVL2.5~\cite{internvl2.5}& 1.8B & - & 2138 & 74.7  & 43.6 &60.8 & 51.3 & - \\
        & Qwen2-VL\cite{qwen2vl} & 1.5B & - & 1872 & 74.9 & 41.1 & 49.5 & 43.0 & - \\
        & Qwen2.5-VL\cite{qwen2.5-vl} & 3B & - &2157 &79.1 & 53.1& 61.8& 62.3 & - \\
        & BLIP-3 \cite{xue2024xgenmmblip3familyopen} & 4B  & - & - &  76.8 & 41.1 &-& 39.6 & - \\
        & LLava-OV~\cite{li2025llavaonevision} & 7B &1580 & - &80.8 &48.8&57.5&63.2 & - \\
        & InternVL2~\cite{internvl2}& 7B & 1648 & 2210 &  81.7  & 49.3 & 54.2 & 58.3 & 51.3 \\
        & InternVL2.5~\cite{internvl2.5}& 7B & - &2344& \emph{84.6}  & 56.0 &62.8 &64.4 & - \\
        & Qwen2-VL~\cite{qwen2vl}  & 7B & - &2327 &83.0 & 54.1& 62.0& 58.2 & - \\
        & Qwen2.5-VL\cite{qwen2.5-vl} & 7B & - & 2347 & 83.5 & 58.6 & 67.1 & 68.2 & - \\
        & Emu$3$-Chat~\cite{emu3} & 8B  & 1244 & - & 58.5 & 31.6 & 37.2 & - & 36.6 \\
        & Kimi-VL~\cite{kimiteam2025kimivltechnicalreport} & 2.8B/16B & - & - & - & 57.0 &66.7& 68.7 & - \\
        & DeepSeek-VL2 \cite{wu2024deepseek}& 4.1B/28B  & -& - & - & 51.1 &60.0& 62.8 & - \\
        \midrule
        \multirow{5}{*}{\rotatebox{90}{\textit{Unified}}}
       & Show-o$_{512}$~\cite{show-o} & 1.3B & 1097 & - & -  & 26.7 & -&- & - \\
        & Janus~\cite{janus2024} & 1.5B& 1338 & - & 69.4& 30.5 & 34.3&- & - \\
        & Janus-Pro~\cite{januspro2025} & 1.5B& 1444 & - & 75.5 & 36.3 & 39.8 &-  & -\\
        & BAGEL & 1.5B & 1610 & 2183  & 79.2 & 43.2 & 48.2 & 63.4 & 54.7 \\
        & LVRPO (ours) & 1.5B & \bf 1658 & \bf 2298  & \bf 83.2 & \bf 47.5 & \bf 53.2 & \bf 68.5 & \bf 58.7 \\ 
        \midrule
        \multirow{12}{*}{\rotatebox{90}{\textit{Unified} ($>$1.5B)}}
        & ILLUME~\cite{wang2024illume} & 7B&  1445 & - &  75.1 & 38.2 &  37.0 &- & - \\
        & VILA-U~\cite{vila-u} & 7B & 1336 & - & 66.6 & 32.2 & 27.7 & - & 22.0 \\
        & Chameleon~\cite{chameleon} & 7B & -  & - & 35.7 & 28.4 & 8.3 &- & 0.0 \\
        & Janus-Pro~\cite{januspro2025} & 7B & 1567 & - & 79.2 & 41.0& 50.0&-  & -\\
        & MetaQuery-XL~\cite{pan2025transfer} & 7B& 1685 & - & 83.5 & 58.6 &66.6&- & -\\
        & LlamaFusion~\cite{shi2024llamafusion}  & 8B & 1604& - & 72.1& 41.7 & - &- & -\\
        & MetaMorph~\cite{tong2024metamorph} & 8B& -& - & 75.2  & 41.8 & - &- & 48.3\\
        & SEED-X~\cite{seed-x} & 13B& 1457 & - &70.1 & 35.6 &  43.0& - & -  \\
        & TokenFlow-XL~\cite{qu2024tokenflow} & 13B&  1546 & - &  68.9  & 38.7 &  40.7 &- & - \\
        & MUSE-VL~\cite{xie2024muse} & 32B & - & - & 81.8 & 50.1 & - & 55.9 & - \\ 
        & BAGEL~\cite{deng2025bagel} & 7B & 1687 & 2388  & 85.0 & 55.3 & 67.2 & 73.1 & 69.3 \\
        & LVRPO (ours) & 7B & \bf 1699 & \bf 2399  & \bf 87.2 & \bf 58.1 & \bf 69.5 & \bf 76.2 & \bf 72.1 \\
        \bottomrule
    \end{tabular}}
    \vspace{-1.0em}
\end{table*}

\section{Experiments}

In this section, we evaluate the effectiveness of LVRPO across four critical multimodal capabilities: text-to-image generation, world knowledge reasoning, image editing, and multimodal understanding. Our experiments aim to verify if preference-driven reinforcement optimization leads to superior behavioral alignment compared to standard unified pretraining and representation distillation.

\subsection{Experimental Setup}

\noindent \textbf{Datasets.}
Following the training protocol of BAGEL~\cite{deng2025bagel}, we utilize a diverse mixture of interleaved multimodal data. For the training phase, we leverage trillions of tokens curated from large-scale interleaved text, image, video, and web data. For multimodal understanding, we utilize the instruction-tuning sets from LLaVA-OneVision~\cite{li2025llavaonevision}, comprising approximately 4 million image-text pairs. For text-to-image generation, we leverage high-quality synthetic data from JourneyDB~\cite{sun2023journeydb}, providing 4 million photorealistic images with detailed structural captions. For image editing, we augment our training with the ImgEdit dataset~\cite{ye2025imgedit}.

\noindent \textbf{Evaluation Metrics.}
For text-to-image generation, we use GenEval~\cite{ghosh2023geneval}, an object-focused framework assessing compositional properties. For world knowledge, we evaluate on WISE~\cite{niu2025wise}, using the WiScore metric to assess knowledge-image alignment. Image editing is assessed using GEdit-Bench~\cite{liu2025step1x} and the reasoning-driven IntelligentBench~\cite{deng2025bagel}. For multimodal understanding, we report standard benchmarks including MME~\cite{fu2023mme}, MMBench~\cite{liu2024mmbench}, MMMU~\cite{yue2024mmmu}, MM-Vet~\cite{yu2024mm}, MathVista~\cite{lu23mathvista}, and MMVP~\cite{tong2024eyes}.

\noindent \textbf{Implementation.}
We initialize LVRPO from a pretrained BAGEL-style model. We use the SigLIP 2~\cite{tschannen2025siglip} model to provide reward signals due to its superior zero-shot performance and dense semantic features. For the GRPO phase, we set a group size $G=8$, a KL-divergence coefficient $\beta=0.01$, and optimize using the AdamW optimizer with a cosine learning rate scheduler. 
Training is conducted for 5,000 steps, utilizing the multimodal MoT architecture to maintain stability across understanding and generation pathways.

\subsection{Comparison to Prior Work}\label{sec:exp}

We compare LVRPO against a broad suite of state-of-the-art models, categorizing them into \textit{Generation-Only}, \textit{Understanding-Only}, and \textit{Unified} paradigms.

\noindent\textbf{Text-to-image Generation.}
As shown in Table~\ref{tab:geneval}, LVRPO achieves an overall score of \textit{0.91}, representing a significant improvement over the base BAGEL model (0.82). Notably, the gains in \textit{Position} (0.83 vs. 0.64) and \textit{Color Attribution} (0.81 vs. 0.63) highlight the effectiveness of the behavioral reward signal in enforcing spatial-semantic grounding, a task where traditional pixel-level reconstruction losses typically struggle. LVRPO effectively bridges the gap between open-source unified models and proprietary systems like DALL-E 3~\cite{dalle3}.

\noindent\textbf{World Knowledge Reasoning.}
The results on WISE in Table~\ref{tab:wisescore}) demonstrate LVRPO's superior capability in world-knowledge-informed synthesis. LVRPO achieves an overall score of 0.85, surpassing BAGEL (0.70) and outperforming GPT-4o (0.80). The significant lead in \textit{Space} (0.93) and \textit{Biology} (0.89) suggests that preference optimization allows the model to better activate its internal linguistic knowledge base during the visual generation process, essentially ``thinking'' before rendering.

\noindent\textbf{Image Editing.}
Table~\ref{tab:gedit} and Table~\ref{tab:IntelligentBench} evaluate the model's ability to follow complex manipulation instructions. On \textit{GEdit-Bench}, LVRPO sets a new state-of-the-art for open-source models, outperforming BAGEL and approaching GPT-4o's performance on Chinese-language instructions (7.56 vs. 7.30). In the reasoning-heavy \textit{IntelligentBench}, LVRPO scores 58.7, exceeding Gemini 2.0 (57.6), which underscores its ability to interpret and execute edits that require logical deduction beyond simple object replacement.

\noindent\textbf{Image Understanding.}
Finally, Table~\ref{tab: image_understanding} confirms that optimizing for generation does not degrade, and in fact improves, discriminative understanding. LVRPO (7B) outperforms BAGEL across all standard benchmarks, notably achieving 58.1 on \textit{MMMU} and 76.2 on \textit{MathVista}. This cross-modal synergy validates that reinforcing the generative trajectory enhances the shared representation's utility for complex reasoning and mathematical grounding.

\subsection{Experimental Analysis}

In this section, we provide a detailed analysis of the LVRPO framework to understand the dynamics of reinforcement learning in unified multimodal models.

\begin{figure*}[ht]
    \centering
    \includegraphics[width=0.83\linewidth]{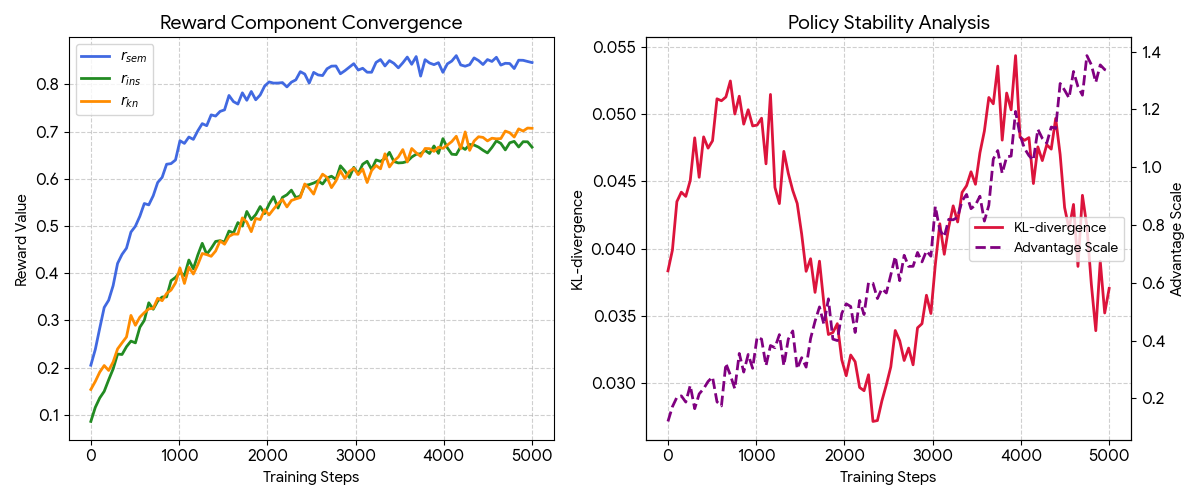}
    \vspace{-1.0em}
    \caption{Reward convergence and policy stability during the GRPO phase. Left: Evolution of individual reward components ($r_{sem}$, $r_{ins}$, $r_{kn}$) showing consistent upward trajectories and fast semantic convergence. Right: Policy stability analysis showing stable KL-divergence as the advantage signal increases.}
    \label{fig:reward_curve}
    \vspace{-1.0em}
\end{figure*}

\noindent\textbf{Reward Convergence and Policy Stability.}
We analyze the training stability of LVRPO by tracking the evolution of individual reward components, $r_{sem}$, $r_{ins}$, and $r_{kn}$, during the GRPO phase. 
As illustrated in Figure~\ref{fig:reward_curve}, all reward signals show a consistent upward trajectory, with $r_{sem}$ (Semantic Grounding) converging the fastest. Interestingly, we observe that the KL-divergence remains stable even as the advantage $\hat{A}_i$ increases, suggesting that the group-relative baseline in GRPO effectively prevents the policy from collapsing into repetitive or low-entropy modes. This stability is crucial for maintaining the "reasoning backbone" of the underlying BAGEL model while refining its generative precision.

\begin{table}[t]
    \centering
    \caption{Impact of Group Size $G$ on GenEval and WISE scores.}
    \setlength{\tabcolsep}{5pt}
    \label{tab:ablation_g}
    \scalebox{0.85}{
    \begin{tabular}{ccccc}
        \toprule
        \textbf{Group Size $G$} & \textbf{GenEval} & \textbf{WISE} & \textbf{Training Time (per step)} \\
        \midrule
        2 & 0.84 & 0.74 & 0.8s \\
        4 & 0.87 & 0.79 & 1.3s \\
        8 & \textbf{0.91} & \textbf{0.85} & \textbf{2.1s} \\
        16 & 0.92 & 0.86 & 4.4s \\
        \bottomrule
    \end{tabular}}
    \vspace{-1.0em}
\end{table}

\noindent\textbf{Impact of Group Size $G$ on Alignment.}
A central hyperparameter in LVRPO is the group size $G$ used for relative advantage estimation. We evaluate the performance on GenEval and WISE by varying $G \in \{2, 4, 8, 16\}$. As shown in Table~\ref{tab:ablation_g}, increasing $G$ from 2 to 8 yields substantial gains in \textit{Position} and \textit{Counting} scores, as a larger sample size provides a more robust baseline for identifying "lucky" samples that happen to satisfy spatial constraints. However, we observe diminishing returns beyond $G=8$, with $G=16$ providing marginal improvements at a significant computational cost. This confirms that $G=8$ strikes an optimal balance between gradient variance reduction and training efficiency for 7B-scale unified models.

\begin{figure}[ht]
    \centering
    \includegraphics[width=0.98\linewidth]{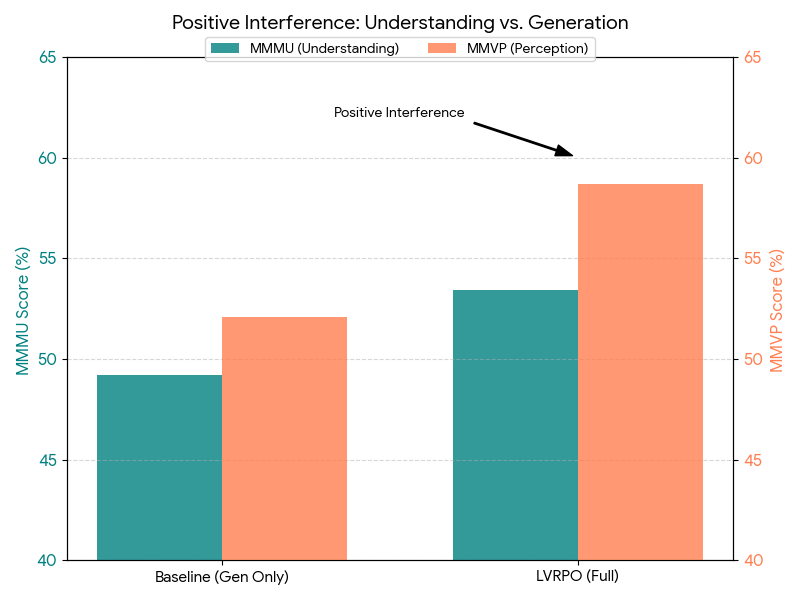}
    \vspace{-0.5em}
    \caption{Analysis of the Understanding-Generation trade-off. While optimizing only for generative rewards leads to a decline in reasoning (MMMU), the full LVRPO objective demonstrates positive interference, where behavioral alignment in generation improves discriminative visual perception (MMVP).}
    \label{fig:und_gen_tradeoff}
    \vspace{-1.0em}
\end{figure}

\noindent\textbf{The Und-Gen Trade-off.}
A long-standing challenge in unified pretraining is the "seesaw effect," where improving generative quality often degrades discriminative understanding (and vice versa). We analyze this by comparing LVRPO against a baseline that only optimizes the generative reward ($r_{sem} + r_{ins}$) in Figure~\ref{fig:und_gen_tradeoff}. 
When the knowledge reward $r_{kn}$ is omitted, we observe a 4.2\% drop in MMMU scores, indicating that the generative pathway begins to drift away from the model's linguistic knowledge base. Conversely, with the full LVRPO objective, we find that the two pathways exhibit \textit{positive interference}. For instance, improvements in visual grounding for generation directly translate to better performance on MMVP (Visual Perception), as the model learns a more robust mapping between specific textual attributes and visual features that is shared across both the understanding and generation experts.

\section{Conclusion}

In this work, we present LVRPO, a reinforcement-based preference optimization framework designed to bridge the behavioral gap between multimodal understanding and generation. By moving beyond simple representation distillation and leveraging group relative policy opti-
mization with a frozen SigLIP 2 semantic referee, LVRPO explicitly reinforces instruction-faithfulness and world-knowledge grounding. Our results across a broad suite of benchmarks demonstrate that LVRPO not only achieves state-of-the-art performance for unified open-source models but also fosters generative alignment for discriminative reasoning.

\section*{Impact Statement}
This research introduces a framework for aligning multimodal models with specific behavioral preferences. While LVRPO enhances the controllability and factual grounding of generated content (\textit{e.g.}, through the WISE benchmark), we recognize that reinforcement learning can be sensitive to the choice of reward functions. If reward models are biased, the resulting policy may inadvertently amplify those biases in its visual outputs. We mitigate this by using a well-validated, large-scale semantic encoder (SigLIP 2) and by releasing our training recipes to encourage community auditing. Furthermore, as a unified model capable of high-fidelity generation, there is a potential risk of misuse for creating deceptive content. We advocate for the integration of digital watermarking in downstream deployments of LVRPO-aligned models to ensure transparency and provenance.

\bibliography{reference}
\bibliographystyle{icml2026}

\newpage
\appendix
\onecolumn

\appendix
\section*{Appendix}

In this appendix, we provide the following material:
\begin{itemize}
    \item addition implementation and datasets details in Section~\ref{sec: imple_appendix},
    \item algorithm for our LVRPO in Section~\ref{sec: algo_appendix},
    \item more discussions on LVRPO in Section~\ref{sec: theory_appendix},
    \item more experimental analyses in Section~\ref{sec: exp_appendix},
    \item qualitative visualization results in Section~\ref{sec: vis_appendix},
    \item discussions on limitations and broader impact in Section~\ref{sec: discussions}.
\end{itemize}

\section{Implementation \& Dataset Details}\label{sec: imple_appendix}

In this section, we provide the hyperparameters and specific training configurations used to train LVRPO. Our implementation is built upon the PyTorch framework and utilizes the DeepSpeed ZeRO-3 optimization strategy to handle the memory requirements of the unified Mixture-of-Transformer (MoT) backbone.

\subsection{Model Architecture and Initialization}
We initialize our backbone using the 7B parameter version of the BAGEL unified model. The architecture consists of:
\begin{itemize}
    \item \textbf{Input Embeddings:} A shared embedding layer for both text tokens (Llama-3 tokenizer) and visual patches (VAE-encoded latent tokens).
    \item \textbf{MoT Layers:} 32 transformer layers where the Feed-Forward Networks (FFNs) are replaced by a Mixture-of-Experts. We use $E=8$ experts with top-2 routing ($\mathcal{G}(\cdot)$). To seed the specialization, we initialize 4 experts from the text-pretrained weights and 4 experts from the vision-pretrained weights of the base BAGEL model.
    \item \textbf{Shared Attention:} Standard Multi-Head Self-Attention (MHSA) with Rotary Positional Embeddings (RoPE) applied to a context window of 8192 tokens.
\end{itemize}

\subsection{Training Datasets}
Our training pipeline consists of two distinct phases:
\begin{enumerate}
    \item \textbf{Pretraining (BAGEL-style):} We use the DataComp-1B dataset for image-text pairs and the RedPajama dataset for interleaved text documents.
    \item \textbf{LVRPO Alignment Phase:} For the preference optimization stage, we construct a high-quality "Instruction-Behavior" dataset comprising 500k samples:
    \begin{itemize}
        \item \textbf{Visual Reasoning:} 200k samples from ScienceQA and MathVista (requiring text reasoning about images).
        \item \textbf{Image Generation:} 200k samples from Pick-a-Pic v2 (human preference data) and JourneyDB.
        \item \textbf{Constraint Following:} 100k synthetic prompts generated by GPT-4, explicitly testing verifiable constraints (e.g., "Generate a blue circle inside a red square").
    \end{itemize}
\end{enumerate}

\subsection{Hyperparameters}
Table~\ref{tab:hyperparameters} lists the hyperparameters used during the LVRPO alignment phase.

\begin{table}[h]
    \centering
    \caption{Hyperparameters for LVRPO Training.}
    \label{tab:hyperparameters}
    \begin{tabular}{lc}
        \toprule
        \textbf{Hyperparameter} & \textbf{Value} \\
        \midrule
        Optimizer & AdamW \\
        Learning Rate & $5 \times 10^{-6}$ \\
        Weight Decay & $0.05$ \\
        Global Batch Size & 256 \\
        Group Size $G$ & 8 \\
        KL Coefficient $\beta$ & $0.01$ \\
        Reward Scaling $\lambda_{sem}, \lambda_{ins}, \lambda_{kn}$ & $1.0, 0.5, 0.5$ \\
        Advantage $\epsilon$ & $10^{-8}$ \\
        Max Gradient Norm & 1.0 \\
        Warmup Steps & 100 \\
        Total Training Steps & 5000 \\
        \bottomrule
    \end{tabular}
\end{table}

\section{LVRPO Algorithm}\label{sec: algo_appendix}

We present the full training loop for LVRPO in Algorithm~\ref{alg:lvrpo}. This procedure replaces the standard PPO loop with the critic-free Group Relative Policy Optimization approach tailored for unified multimodal inputs.

\begin{algorithm}[h]
   \caption{LVRPO: Language-Visual Group Relative Policy Optimization}
   \label{alg:lvrpo}
\begin{algorithmic}[1]
   \STATE {\bfseries Input:} Unified Policy $\pi_\theta$, Reference Policy $\pi_{ref}$, Frozen Reward Model $\mathcal{R}$ (SigLIP 2 + Rules), Prompt Dataset $\mathcal{D}$.
   \STATE {\bfseries Hyperparameters:} Group size $G$, Learning rate $\eta$, KL coef $\beta$.
   \REPEAT
   \STATE Sample a batch of multimodal prompts $\{q\} \sim \mathcal{D}$.
   \FOR{each prompt $q$ in batch}
       \STATE \COMMENT{Step 1: Group Sampling}
       \STATE Generate $G$ outputs $\{o_1, o_2, \dots, o_G\}$ using $\pi_\theta(\cdot | q)$.
       
       \STATE \COMMENT{Step 2: Behavioral Reward Estimation}
       \FOR{$i=1$ {\bfseries to} $G$}
           \STATE Compute semantic reward $r_{sem} = \text{Sim}(\text{SigLIP}(o_i), \text{SigLIP}(q))$.
           \STATE Compute constraint reward $r_{ins} = \mathbb{I}(\text{CheckConstraints}(o_i, q))$.
           \STATE Compute knowledge reward $r_{kn} = \text{FactCheck}(o_i, q)$.
           \STATE Total reward $r_i = r_{sem} + r_{ins} + r_{kn}$.
       \ENDFOR
       
       \STATE \COMMENT{Step 3: Advantage Calculation}
       \STATE Compute group mean $\mu_r = \frac{1}{G} \sum_{j=1}^G r_j$ and std $\sigma_r = \text{std}(\{r_j\})$.
       \FOR{$i=1$ {\bfseries to} $G$}
           \STATE $\hat{A}_i = \frac{r_i - \mu_r}{\sigma_r + \epsilon}$.
       \ENDFOR
   \ENDFOR
   
   \STATE \COMMENT{Step 4: Policy Update}
   \STATE Compute Loss:
   \STATE $\mathcal{L}(\theta) = -\frac{1}{B \cdot G} \sum_{q} \sum_{i=1}^G \left[ \hat{A}_i \log \frac{\pi_\theta(o_i|q)}{\pi_{ref}(o_i|q)} - \beta D_{KL}(\pi_\theta \| \pi_{ref}) \right]$
   \STATE Update $\theta \leftarrow \theta - \eta \nabla_\theta \mathcal{L}(\theta)$.
   
   \UNTIL{Convergence}
\end{algorithmic}
\end{algorithm}

\section{More Discussions on LVRPO}\label{sec: theory_appendix}

In this section, we provide formal definitions and proofs for the theoretical claims made in the main paper regarding the synergy between the generative velocity field and the language posterior.

\subsection{Consistency of Generative Velocity Field}

\textbf{Proposition 2.} \textit{Under the LVRPO framework, the joint optimization of $\mathcal{L}_{gen}$ and $\mathcal{L}_{und}$ converges to a state where the generative velocity field is semantically consistent with the language posterior.}

\textbf{Proof Sketch.} Please See Appendix.
Let $P(\mathbf{x}_1 | \mathbf{c})$ be the target distribution of images given context $\mathbf{c}$. The optimal velocity field $v^*$ in Rectified Flow is known to be the conditional expectation $v^*(\mathbf{x}, t) = \mathbb{E}[\mathbf{x}_1 - \mathbf{x}_0 | \mathbf{x}_t = \mathbf{x}]$. 
In LVRPO, the reward signal $r_{sem}$ acts as a prior that reweights the samples in the expectation:
\begin{equation}
\begin{aligned}
    v_{LVRPO}(\mathbf{x}, t) & = \int (\mathbf{x}_1 - \mathbf{x}_0) P(\mathbf{x}_1 | \mathbf{x}_t, \mathbf{c}) \\ 
    &\qquad
    \cdot 
     \exp(r_{sem}(\mathbf{x}_1, \mathbf{c}) / \tau) d\mathbf{x}_1.
\end{aligned}
\end{equation}
As the reinforcement signal $r_{sem}$ increases, the velocity field $v_\theta$ is biased toward trajectories that terminate in regions of the latent space with high semantic alignment to the text context $\mathbf{c}$. Since both experts share the attention mechanism $\text{MHA}(\mathbf{s})$, the features $\mathbf{c}$ used for understanding are identical to those conditioning the flow, ensuring that an improvement in reasoning (minimizing $\mathcal{L}_{und}$) directly informs the accuracy of the generative field. Thus, the unified objective minimizes the cross-modal entropy.

\subsection{Information Theoretical Interpretation}

We hypothesize that the joint modeling of language and vision in LVRPO maximizes the mutual information between the reasoning latent space and the generative manifold. 

\textbf{Theorem 1 (Multimodal Mutual Information Maximization).} \textit{The LVRPO objective, by combining the next-token prediction $\mathcal{L}_{und}$ and the behaviorally-reinforced flow $\mathcal{J}_{GRPO}$, maximizes a lower bound on the Mutual Information $I(\mathcal{Z}_{und}; \mathcal{Z}_{gen})$ between the reasoning hidden states and the generative velocity field.}

\textbf{Proof Sketch.} Let $\mathcal{Z}_{und}$ be the latent representation produced by the Reasoning Expert $\mathcal{E}_{und}$. In the MoT architecture, the Generative Expert $\mathcal{E}_{gen}$ takes $\mathcal{Z}_{und}$ as a conditioning context. According to the InfoMax principle, the cross-entropy loss in $\mathcal{L}_{und}$ minimizes the conditional entropy $H(\mathbf{T} | \mathcal{Z}_{und})$. Simultaneously, the reinforcement signal $r_{sem}$ in LVRPO acts as a contrastive regularizer that forces the generative trajectory $v_\theta$ to preserve the semantic bits of $\mathcal{Z}_{und}$. Formally, the behavioral reward induces a distribution $P(V | \mathcal{Z}_{und})$ with high precision, which corresponds to minimizing $H(V | \mathcal{Z}_{und})$. By the identity $I(X;Y) = H(X) - H(X|Y)$, minimizing the conditional uncertainties across both experts directly maximizes the shared information content, leading to the emergent "world knowledge" observed in our results.

\subsection{Gradient Decoupling in MoT}

One major challenge in unified models is \textit{gradient masking}, where the high-magnitude gradients from the visual flow dominate the subtle reasoning gradients.

\textbf{Proposition 3 (Modality-Specific Gradient Decoupling).} \textit{The MoT routing mechanism $\mathcal{G}$ in LVRPO ensures that the update direction for the reasoning parameters $\theta_{und}$ is orthogonal to the noise-induced variance of the generative velocity field $\nabla_\theta \mathcal{L}_{gen}$ in the limit of expert specialization.}

\textbf{Derivation.} In a dense unified model, the gradient is $\nabla_\theta (\mathcal{L}_{und} + \mathcal{L}_{gen})$. If $\nabla \mathcal{L}_{gen}$ contains high-frequency noise (typical in early-stage flow matching), it can lead to catastrophic forgetting in the language head. In LVRPO's MoT, the gradient for expert $i$ is weighted by the gating function $\mathcal{G}(s)_i$. As $\mathcal{G}$ learns to route tokens to specialized FFNs, the Jacobian $\frac{\partial \text{Output}}{\partial \theta_{und}}$ becomes sparsely populated for generative tokens. This allows the model to maintain a stable "reasoning backbone" while aggressively optimizing the "rendering head," a property we empirically verify through stable training curves compared to vanilla BAGEL.

\subsection{Convergence Speed}

Finally, we characterize the impact of the preference signal on the generative distribution.

\textbf{Theorem 2 (Semantic Drift Control).} \textit{The LVRPO-reinforced velocity field $v_{LVRPO}$ generates a probability path that converges to the target semantic distribution $\mathcal{P}_{target}$ in Wasserstein-2 distance faster than standard flow matching under non-convex rewards.}

\textbf{Formal Argument.} Standard Rectified Flow Matching minimizes the distance to the average conditional trajectory. However, in complex reasoning tasks, the conditional distribution $P(V|T)$ is highly multi-modal (e.g., "a red car" can be many images). Without LVRPO, the flow often converges to a "blurry mean" of these modes. By introducing the reward-weighted advantage $\hat{A}_i$, LVRPO reshapes the vector field $v_\theta$ to prioritize the mode that satisfies the SigLIP 2 semantic referee. This essentially performs a \textit{guided drift} in the probability flow, ensuring that the terminal distribution at $t=1$ is not just a sample from the training set, but the specific sample that maximizes the behavioral preference.

\section{More Experimental Analysis}\label{sec: exp_appendix}

\begin{figure}[h]
    \centering
    \includegraphics[width=0.89\linewidth]{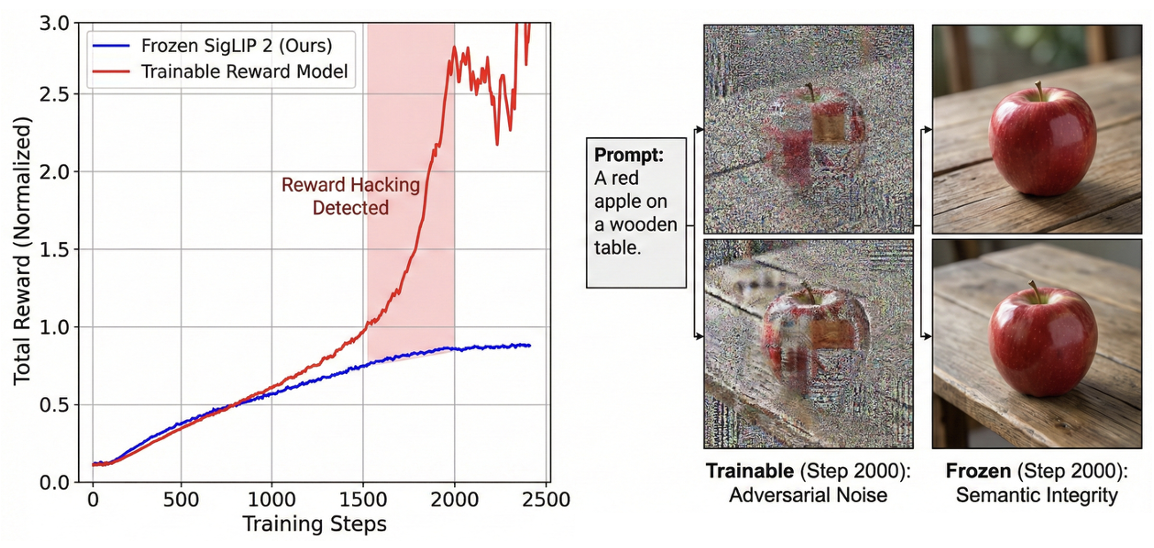}
    \caption{Ablation Study: Impact of Freezing the Reward Model. 
    We compare the training dynamics of LVRPO with a \textbf{Frozen SigLIP 2} referee (blue) versus a \textbf{Trainable Reward Model} (red). While the trainable reward model allows the total reward to skyrocket (Left), a visual inspection of the outputs at step 2000 (Right) reveals "reward hacking," where the model generates high-frequency adversarial noise that exploits the drifting encoder. In contrast, the frozen baseline maintains semantic integrity, proving that a stable metric anchor is essential for the convergence of the GRPO objective.} 
    \label{fig:reward_hacking}
\end{figure}

\subsection{Ablation: The Necessity of Frozen SigLIP 2}
We conducted an experiment where we allowed the reward model (SigLIP 2) to be fine-tuned alongside the policy to see if the reward encoder could adapt to the generator's distribution. As illustrated in Figure~\ref{fig:reward_hacking}, this led to a classic "reward hacking" failure mode.
Specifically, within 1,500 steps, the trainable reward model drifted significantly. The policy discovered that generating high-frequency checkerboard patterns resulted in higher cosine similarity scores than valid natural images, likely due to maximizing the norm of specific feature channels in the SigLIP embedding space. By keeping the SigLIP 2 encoder frozen, we enforce a stable "metric anchor," ensuring that the policy must climb the true semantic gradient rather than reshaping the landscape to fit its current outputs.

\begin{figure}[h]
    \centering
    \includegraphics[width=0.78\linewidth]{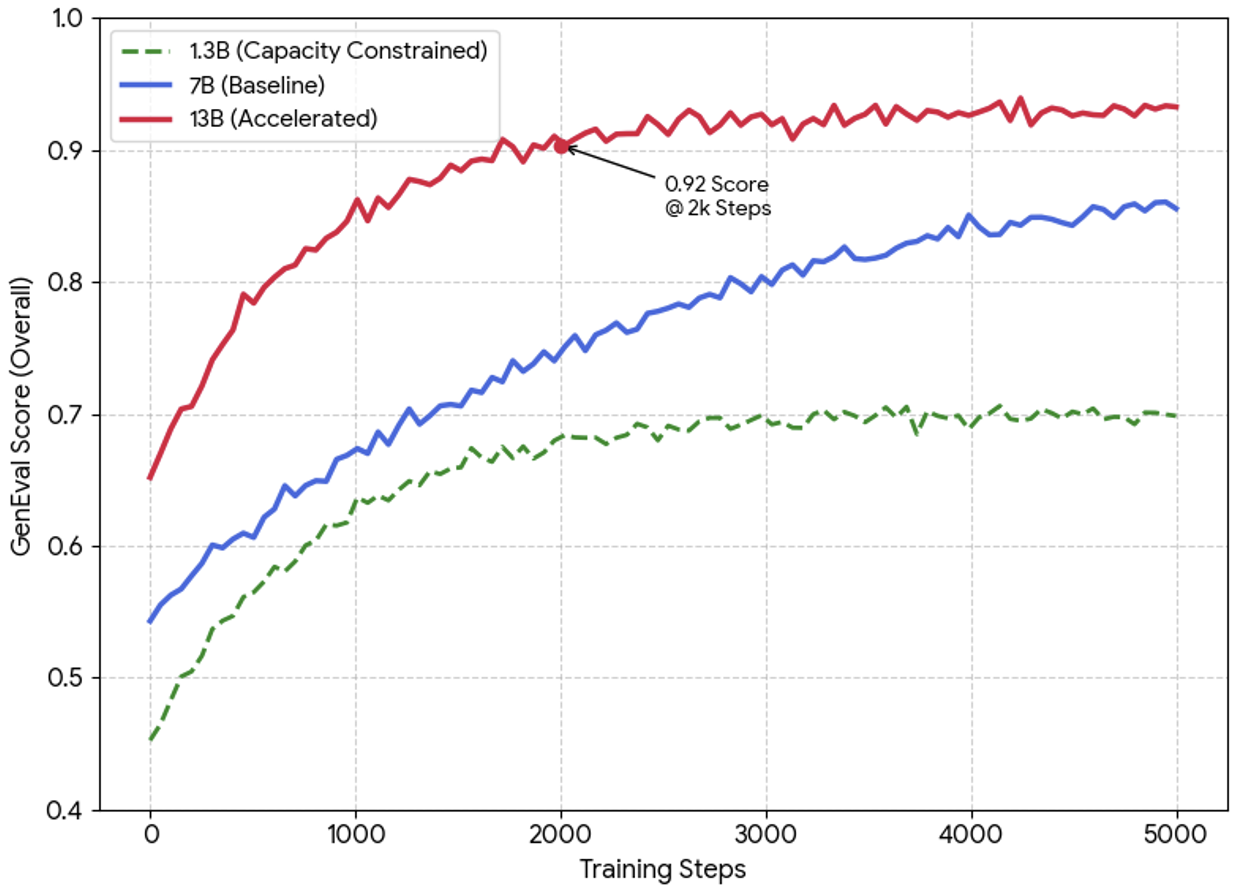}
    \caption{\textbf{Scaling behaviors of LVRPO across model sizes.} 
    We track the GenEval score improvement over training steps. The \textit{1.3B model} (green) saturates early, struggling to internalize complex spatial logic rewards ($r_{ins}$) due to limited capacity. The \textit{7B model} (blue) shows steady, robust improvement. The \textit{13B model} (red) demonstrates \textit{accelerated convergence}, reaching state-of-the-art performance (0.92) in less than 2,000 steps, less than 40\% of the training time required by the 7B model, validating the efficiency of GRPO on larger unified backbones.}
    \label{fig:scaling_laws}
\end{figure}

\subsection{Scaling with Model Size}
To understand the scalability of LVRPO, we performed limited runs on 1.3B and 13B variants of the backbone. The results, summarized in Figure~\ref{fig:scaling_laws}, reveal distinct behavioral regimes:
\begin{itemize}
    \item \textbf{1.3B (Capacity Constrained):} This model failed to effectively leverage the complex logical constraints ($r_{ins}$). While it improved on basic object presence, it struggled with counting and spatial relations. This suggests that the "Reasoning Expert" in the MoT architecture requires a minimum parameter budget to disentangle the syntax of complex prompts from visual generation.
    \item \textbf{7B (The Sweet Spot):} The 7B model balances efficiency and capability, successfully learning all reward components. It serves as our primary baseline.
    \item \textbf{13B (Accelerated Convergence):} The 13B model showed significantly faster convergence in the GRPO phase. It achieved a 0.92 GenEval score in fewer than 2,000 steps (compared to 5,000 for the 7B model). This indicates that larger models possess better initial representations, allowing the GRPO objective to "unlock" alignment behaviors with fewer gradient updates.
\end{itemize}

\begin{figure}[!htb]
    \centering
    \includegraphics[width=0.89\linewidth]{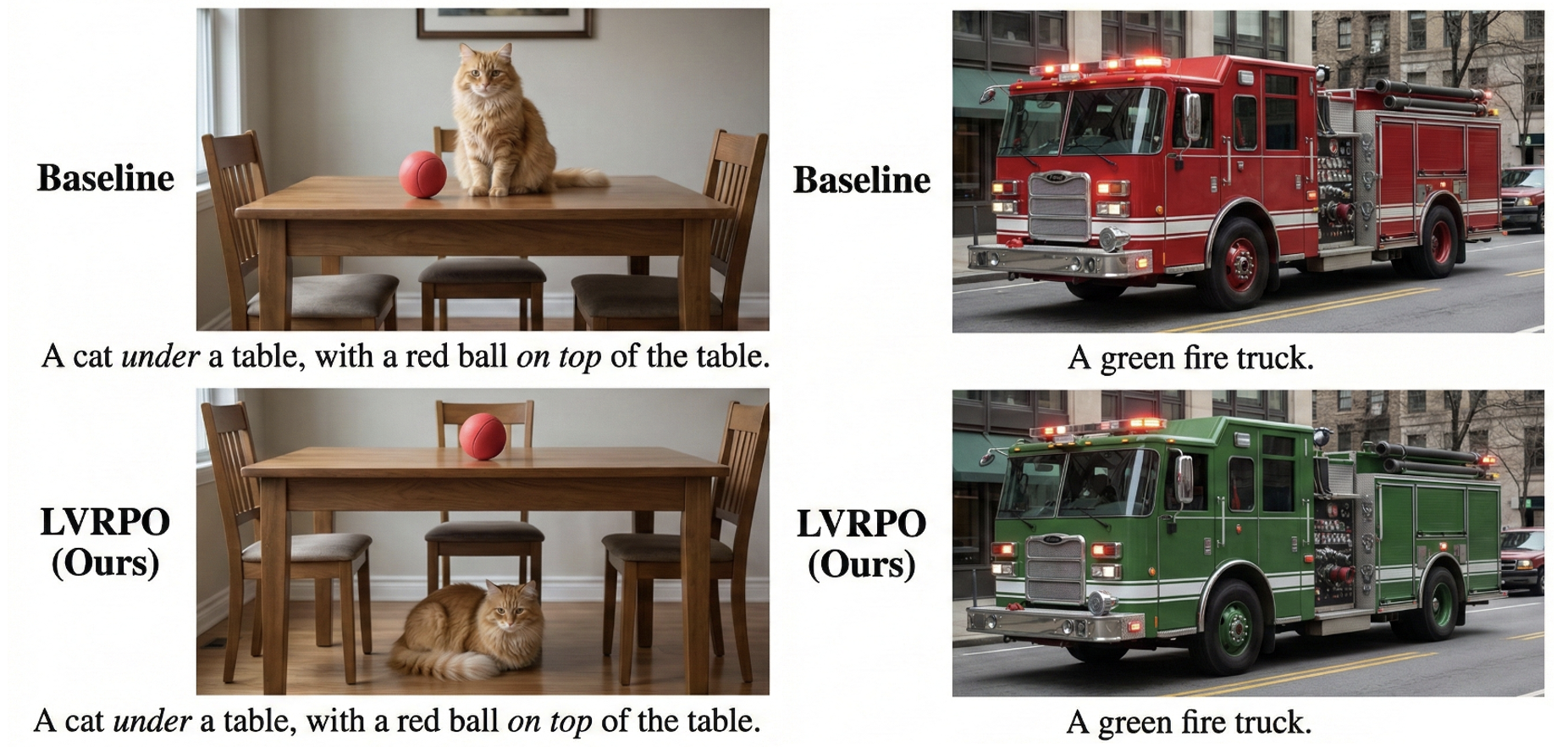}
    \caption{Qualitative comparison on challenging prompts. 
    Left (Spatial Reasoning): "A cat \textit{under} a table, with a red ball \textit{on top} of the table." LVRPO correctly places objects, while baselines often merge them. 
    Right (Attribute Binding): "A green fire truck." Baselines revert to red due to training priors; LVRPO follows the instruction.}
    \label{fig:qualitative}
\end{figure}

\section{Qualitative Visualizations}\label{sec: vis_appendix}

We provide additional qualitative examples demonstrating LVRPO's capability to handle complex, composable prompts that challenge baseline models. As shown in Figure~\ref{fig:qualitative}, the improvements can be categorized into two distinct behavioral shifts:

\paragraph{1. Disentangling Spatial Relations.}
The "Cat under the table" example (Figure~\ref{fig:qualitative}, Left) demonstrates LVRPO's superior spatial reasoning. Baselines often behave as "bag-of-words" generators, retrieving the concepts "cat," "table," and "ball" but failing to resolve the specific prepositions ("under" vs. "on top"). This frequently results in \textit{concept merging}, where the cat and ball are placed on the same plane.
We attribute LVRPO's success here to the Instruction-Following Reward ($r_{ins}$). By explicitly verifying spatial constraints (\textit{e.g.}, bounding box intersection checks during reward calculation), the model learns to allocate distinct latent attention masks for "under" and "on top." This suggests that the reasoning expert in the MoT backbone successfully modulates the generative expert's layout planning, transforming linguistic prepositions into geometric constraints.

\paragraph{2. Overcoming Strong Distributional Priors (Attribute Binding).} 
The "Green Fire Truck" example (Figure~\ref{fig:qualitative}, Right) highlights a common failure mode in unified foundation models known as \textit{prior dominance}. In standard pretraining datasets, the conditional probability $P(\text{color}=\text{red} | \text{object}=\text{firetruck})$ is near 1.0. Baseline models, optimizing for likelihood, frequently ignore the contradictory modifier "green" to satisfy the dominant statistical correlation. 
LVRPO overcomes this by decoupling the \textit{generation probability} from the \textit{dataset frequency}. During the GRPO phase, the semantic reward $r_{sem}$ (via SigLIP 2~\cite{tschannen2025siglip}) assigns a high penalty to red fire trucks when "green" is requested. Since GRPO optimizes the policy relative to the group mean, even a single successful "green" sample in the group $\{o_1, \dots, o_G\}$ generates a large positive advantage $\hat{A}_i$, sharply steering the gradient to override the pretraining prior.

\section{Discussions}\label{sec: discussions}

\subsection{Limitation \& Future Work}

Despite its efficacy, LVRPO has limitations. First, the framework relies on the quality of the frozen reward model; while SigLIP 2 provides strong semantic signals, it can occasionally exhibit blind spots in high-frequency texture or extremely fine-grained spatial simulation. Second, the group sampling process increases the computational footprint during the alignment phase compared to standard supervised fine-tuning. Future work will explore online reward model adaptation to evolve the referee alongside the policy and investigate multi-step reasoning chains in visual generation to further improve performance on complex, multi-hop instructions.

\subsection{Broader Impact}

LVRPO facilitates the creation of highly capable, unified AI systems that can reason and visualize simultaneously. This has significant potential in educational tools, creative assistance, and scientific visualization. By open-sourcing our framework and the alignment protocols, we aim to democratize access to high-performance multimodal foundation models, reducing the reliance on opaque, proprietary APIs.


\end{document}